\documentclass[10pt]{article}

\usepackage[margin=1in]{geometry}
\usepackage{lmodern}
\usepackage[T1]{fontenc}
\usepackage{amsmath,amssymb,amsthm}
\usepackage{booktabs}
\usepackage{microtype}
\usepackage{textcomp}
\usepackage{xcolor}
\usepackage{tikz}
\usetikzlibrary{positioning,arrows.meta,fit,calc,shapes.misc}
\usepackage[colorlinks=true,linkcolor=blue!60!black,citecolor=blue!60!black,urlcolor=blue!60!black]{hyperref}

\newtheorem{definition}{Definition}
\newtheorem{proposition}{Proposition}

\newtheorem{remark}{Remark}

\newcommand{\us}{\,\textmu s}
\newcommand{\ms}{\,ms}
\newcommand{\Prefix}{\mathrm{Prefix}}
\newcommand{\grid}{GRID}
\newcommand{\code}[1]{\texttt{#1}}

\title{\textbf{GRID: Grammar-Railed Decoding for Enterprise SQL Generation}%
\thanks{A condensed version of this paper is under review at KDD 2027.}}

\author{Mohsen Arjmandi\\
\small evolutionID GmbH\\
\small \texttt{mohsen.arjmandi@evolutionid.com}}

\date{July 2026}

\begin{document}
\maketitle

\begin{abstract}
Large language models can write SQL, but enterprise deployment demands more than
plausible text: outputs must be syntactically valid, must respect per-role and
per-schema policy, must come with provable---not best-effort---guarantees, must not
slow down as generations grow, and must leave a compliance-grade record of every
decision. We present \grid{} (Grammar-Railed Decoding), a grammar-constrained
decoding engine that keys exact next-token masks on \emph{parser configurations}
(lexer scan state $\times$ LALR(1) stack) rather than on token sequences, uses the
incrementally advanced LALR(1) parser itself as a viable-prefix oracle, and bridges
LLM tokens to grammar terminals through a byte-level trie walk with a
context-independent/context-dependent split that makes cache-key soundness hold by
construction. Role-based access control is compiled \emph{into the language}: role
projections subset the grammar's productions and schema lexicons restrict identifier
terminals, so forbidden verbs and identifiers are unreachable at mask level. Four
guarantees---soundness, completeness, termination, and near-constant per-token cost
(requirement~R)---are stated with explicit preconditions and each is paired with the
test or benchmark that verifies it. Rust kernels bring the warm serving step to
1.33\us{} per request and the per-token mask to a 3.6--6.7\us{} median---ahead
of llguidance at both p50 and p90 on both benchmarked tokenizers, with zero
false rejects, while llguidance keeps the flattest p99. On a declared H100
SXM5 runner, end-to-end serving overhead under vLLM is $+1.51\%$
time-per-output-token at batch~32 with 27.3\ms{} cold-schema specialization; a
fresh (never-seen) schema pays $\sim$0.7\ms{} first-token latency and then
decodes at $1.00\times$ warm speed. A v7 kernel that keeps the entire cold mask
miss in Rust (fused walk$\to$blob$\to$register, one GIL-released call) holds the
warm co-tenants' worst engine step to 15.3\ms{} on real hardware; the one
residual cost---a transient co-batched TPOT slowdown ($\sim$34\%) during the
fresh schema's $\sim$0.66\,s specialization window---is genuine host
CPU/memory-bandwidth contention between the cold walk and the engine forward loop
(it shrinks as walk threads grow), a compute-isolation trade-off noted as future
work rather than a defect. Per-token guard cost is
position-flat (slope $\approx 0$ at $n{=}16{,}000$, all nesting depths). On Spider, constrained decoding is worth $+13$ execution-accuracy points
at 0.5B; at 7B the mask alone is worth $\sim$$+1$ point, and one
checker-guided repair pass over the provably mask-unenforceable residue
(column-level policy) lifts execution to $94.5\%$ and EX to $+2.3$ over
unconstrained. A hash-chained per-token audit trail replays bit-identically
(1{,}000 generations across a cache-namespace rollover; $100\%$ tamper detection).
We state honestly what the mask cannot do---distribution faithfulness, column-level
RBAC, non-LALR(1) languages---and where measured cost remains (cold trie walks;
the serving cold-window compute contention; one characterized engine-side artifact
exogenous to \grid{}).
\end{abstract}

\section{Introduction}\label{sec:intro}

We are concerned with generating SQL (and, more generally, sentences of any
LALR(1)-parsable language) from a large language model under the operating
conditions of an enterprise CRUD/RBAC deployment. In that setting, ``the model
usually writes valid SQL'' is not an acceptable contract. Six problems, each of
which shapes the design, must be solved \emph{simultaneously}:

\begin{enumerate}
\item[\textbf{P1.}] \textbf{Validity and policy, jointly.} Output must parse under
the SQL dialect \emph{and} comply with per-role policy (allowed verbs, banned
clauses) and per-schema vocabulary (only real tables and columns). Policy is not a
post-hoc filter: a role that may only \code{select} must be unable to emit
\code{delete} at all, and identifiers must be schema-valid \emph{by construction}.
\item[\textbf{P2.}] \textbf{Provable guarantees.} Soundness (never emit a token
that leaves the language's prefix set), completeness (never block a token that
could still lead to a valid sentence), and termination (every stop is a complete
sentence) must be theorems with explicit preconditions, each paired with the
test that verifies it---not emergent properties of a test suite.
\item[\textbf{P3.}] \textbf{Flat per-token cost (requirement R).} Guard-rail cost
per token must be amortized $O(1)$ in the current output length $n$ (worst case
bounded by grammar/nesting constants, never by $n$), so total guard cost is $O(n)$
over a generation. A guard whose per-token cost grows with context is unusable for
long generations; \S\ref{sec:eval-r} shows a 2023-era system for which
per-token overhead grows by $\sim$1.19\ms{} \emph{per position}.
\item[\textbf{P4.}] \textbf{A replayable audit trail.} In a compliance setting,
``what was the model allowed to say at step $t$, and why?'' must be answerable
after the fact: a hash-chained, per-token record of permit/block decisions,
replayable bit-identically against versioned grammar artifacts.
\item[\textbf{P5.}] \textbf{Serving reality.} Production decoding is batched,
grammars are heterogeneous across co-batched requests, and new schemas arrive
while a hot batch is running. The guard must overlap mask computation with the GPU
forward pass, never stall co-batched requests on one request's cold miss
(the \emph{cold-schema-into-hot-batch} problem), and never substitute an
approximate mask under deadline pressure.
\item[\textbf{P6.}] \textbf{An honest boundary for the unenforceable residue.}
Some policy is \emph{provably} not enforceable by any left-to-right
context-free mask---column-level RBAC being the canonical case
(Proposition~\ref{prop:column-rbac}). The system must name that boundary rather
than blur it, enforce it post-parse, and---where a capable model is
available---convert the named violations into a checker-guided repair loop.
\end{enumerate}

\grid{} answers these with one architectural idea and its consequences:
\emph{key everything on parser configurations}. A configuration---the pair of the
lexer's scan state over a partial lexeme and the LALR(1) parser stack---is
exactly the information that determines which continuations are viable. Masks are
computed from configurations, cached under configuration-derived keys whose
soundness is a stated obligation (a Myhill--Nerode-style refinement,
\S\ref{sec:keys}), and audited under a rolling configuration hash. The mask cost
therefore tracks the \emph{grammar configuration}, not the output position, which
is requirement~R by construction; the benchmarks in \S\ref{sec:eval} measure that
this holds on real vocabularies at $n = 16{,}384$.

\paragraph{Contributions.}
(1)~A formal framework for configuration-keyed constrained decoding: the
incremental LALR(1) parser as a viable-prefix oracle with exact EOS gating under
table compression (\S\ref{sec:oracle}); a byte-level token$\leftrightarrow$terminal
bridge with a context-independent/context-dependent split that makes cache-key
soundness hold by construction (\S\ref{sec:bridge}); the cache-key soundness
obligation OBL-KEY1 and the identifier composition rule, with the scanner
normal form that maximizes sound cross-request sharing (\S\ref{sec:keys}).
(2)~Four guarantees stated as propositions with explicit preconditions, including
the honest negative result that column-level RBAC is mask-unenforceable
(\S\ref{sec:guarantees}).
(3)~A system realizing the framework---grammar pipeline, Rust kernels
(v4$\to$v7, warm serving step $7.39\to1.33$\us{}/request, then a v7 fused
cold-miss path), two-tier write-back
mask cache, a serving contract for batched heterogeneous decoding, and a
hash-chained audit log (\S\ref{sec:system})---with worked examples
(\S\ref{sec:examples}).
(4)~A committed benchmark record: per-token mask latency against XGrammar,
llguidance, and Outlines on two tokenizers; requirement-R slope measurements;
MaskBench; Spider execution accuracy with checker-guided repair; and an
end-to-end vLLM serving record including a cold-miss stress arm
(\S\ref{sec:eval}).

\paragraph{Design lineage.}
The initial patent filing captured an early, partial snapshot of this design
(v0.0.5, designed August 2023): a precomputed validity table over token sequences
against a pushdown automaton, with decode-time logit masking. Planned iteration
carried the design forward: v0.0.6 (September 2023) already outperformed the
July-2023 release of \code{guidance} on the scaling benchmarks of that era, and
v0.0.7 (November 2023)---the system described in this paper, and the line the
final patent application followed---replaced sequence keys with configuration
keys and acceptance semantics with prefix viability, added the byte-level
token$\leftrightarrow$terminal bridge, the write-back cache, RBAC/schema
projections, the audit chain, and checker-guided repair; the work has continued
to improve since. Sections~\ref{sec:oracle}--\ref{sec:bridge} present the design
analysis behind the two load-bearing revisions. The design drew inspiration from
published work---notably Willard and Louf's Outlines
paper~\cite{willard2023efficient}, whose finite-state-machine formalization of
guided generation this paper deliberately parallels in register---while
\grid{}'s design and implementation are our own throughout.

\section{Preliminaries}\label{sec:prelim}

\paragraph{Masked decoding.}
Let $V$ be a tokenizer vocabulary with $|V| = N$ and let
$\Sigma$ be the byte alphabet. An autoregressive LM defines logits
$\boldsymbol\alpha = \mathrm{LM}(s_{1..t},\boldsymbol\theta) \in \mathbb{R}^N$;
guided generation applies a mask $m \in \{0,1\}^N$ before sampling,
$\tilde{\boldsymbol\alpha} = m \odot \boldsymbol\alpha$ (implemented as an
in-place \code{masked\_fill\_($-\infty$)}), and samples
$s_{t+1} \sim \mathrm{Categorical}(\tilde{\boldsymbol\alpha})$. Constructing $m$
at each step is the entire problem: done naively it costs a parse of the whole
prefix per candidate token per step.

\paragraph{Tokens are byte strings.}
Each token has a canonical spelling
$\mathrm{bytes}\colon V \to \Sigma^\ast$ (\code{token\_bytes}): byte-level-BPE
unicode remaps inverted, SentencePiece/BPE space markers normalized to
\code{0x20}, byte-fallback literals \code{<0xNN>} mapped to their byte. One
definition serves the trie build, the fast path, and the reference oracle.
Distinct token ids may share a spelling (\emph{alias classes}); masks are over
ids, not spellings.

\paragraph{Three grammar layers, one language.}
\begin{itemize}
\item \textbf{L1 (dialect core).} A CFG $G = (\mathcal N, T, P, S)$ whose terminals
$\tau \in T$ carry regular lexeme languages $R_\tau \subseteq \Sigma^\ast$, with an
ignored subset $I \subseteq T$ (whitespace, comments) that is first-class. Production
SQL grammars (PostgreSQL, MySQL, SQLite) are LALR(1), so deterministic-PDA coverage
suffices. Lexing is maximal munch over the union automaton with \emph{contextual
resolution}: a forced emission event carries the full candidate terminal set at the
longest match and the consumer picks the highest-priority \emph{parser-viable}
candidate (this is what lets \code{TABLE\_NAME} and \code{COLUMN\_NAME} share one
regex, and what resolves JSON's property-key-vs-string overlap with no new
mechanism).
\item \textbf{L2 (role projection).} A production subset $P_{\mathrm{role}}
\subseteq P$ (verb subsets, clause bans) followed by \emph{mandatory}
useless-symbol elimination and a verified $L(G_{\mathrm{role}}) \neq \emptyset$.
Reducedness is load-bearing: without it, ``non-empty action set $\Rightarrow$
viable prefix'' fails and dead ends return.
\item \textbf{L3 (schema lexicon).} For identifier categories $C \subseteq T$,
finite allow-lists $W_c \subseteq \Sigma^\ast$ realized as lexer tries; the
lists are generated from the database catalog
(\code{information\_schema}). Precondition, validated at
guide build rather than assumed: $W_c \subseteq R_{c}$ for every category (each
allowed word must be scannable to an accepting state of its terminal).
\end{itemize}
The constrained language $L = L(G_{\mathrm{role}},\mathrm{schema})$ is the set of
byte strings that lex and parse under the projected grammar with every
category-$c$ lexeme in $W_c$. Both the policy bundle and the schema snapshot are
fingerprinted, immutable inputs; grammars, tries, and reserve tables are
content-addressed artifacts.

\section{The viable-prefix oracle}\label{sec:oracle}

\subsection{Design analysis: why sequence keys and acceptance semantics fail}
\label{sec:v005}

The v0.0.5 design snapshot precomputed an in-memory table tagging LLM
\emph{token sequences} as accepted or rejected by a pushdown automaton, and
masked per step from table lookups. Two observations, made during the planned
design review of that snapshot, force the shape of everything that follows.

\begin{proposition}[Sequence-keyed tables scale combinatorially]
\label{prop:seq-keys}
A total validity table over token sequences of length $\le m$ has
$\Theta(|V|^m)$ entries. At $|V| = 32{,}768$ and $m = 3$ this is already
$\approx 3.3\times10^{13}$ entries ($\approx 400$\,TB at 12 bytes/entry); at
$m = 18$ it exceeds the number of atoms in the observable universe.
\end{proposition}

A ``representative subset'' of that table is defined by outcome, not by a
construction, so it offers no build recipe. The information the table was trying
to enumerate is, however, finitely presentable: by the viable-prefix property of
LR automata~\cite{knuth1965translation}, the set of viable prefixes of an LR
grammar is characterized by the parser's configurations, a
\emph{grammar-sized} key space ($10^4$--$10^6$ for SQL-class grammars).

\begin{proposition}[Acceptance semantics deadlock at step 1]
\label{prop:acceptance}
Let the tag semantics be ``the PDA accepts this sequence'' (membership in $L$).
For any $L$ containing no sentence of length one token, every single-token
continuation of the empty prefix is tagged rejected, so the step-1 mask is empty
and generation cannot start. Masking requires $\Prefix(L) = \{w : \exists w',\
w w' \in L\}$, not $L$.
\end{proposition}

Both conclusions are constructive, not merely destructive: they identify the
correct key space (configurations) and the correct semantics (prefix viability),
which the rest of this section formalizes.

\subsection{Configurations and the correct-prefix property}

\begin{definition}[Parser configuration]\label{def:config}
A \emph{configuration} is a pair $\kappa = (\sigma, \rho)$ where $\sigma$ is the
LALR(1) parser stack (a persistent, immutable-node chain of states) and $\rho$ is
the lexer scan state over the current partial lexeme---concretely the remainder
bytes $r$ of the single in-progress lexeme, which determine the scanner-DFA state,
the maximal-munch hypothesis set, and the last-accept position deterministically.
\end{definition}

\begin{definition}[Viable prefix]\label{def:viable}
$w \in \Prefix(L) \iff \exists w' \in \Sigma^\ast : w\,w' \in L$. Operationally:
the incremental scan-and-shift of $w$ succeeds and the trailing partial lexeme is
live for some allowed-or-ignored terminal.
\end{definition}

\begin{proposition}[The incremental parser is the oracle]\label{prop:oracle}
Under the preconditions of \S\ref{sec:prelim} (reduced projected grammar;
validated lexicons $W_c \subseteq R_c$), the incrementally advanced LALR(1)
parser with the maximal-munch contextual lexer accepts exactly $\Prefix(L)$:
by the correct-prefix property~\cite{aho1974lr}, an LR parser detects an error
at the first terminal that cannot extend any sentence, so ``no error so far''
$\equiv$ ``viable prefix.'' No separate recognizer is needed.
\end{proposition}

Two refinements make this exact in practice rather than merely in principle.

\paragraph{Allowed terminals under table compression.}
LALR tables with default reductions over-approximate raw action rows (spurious
reduces). The allowed-terminal set is therefore computed by \emph{simulation}:
$A(\sigma) = \{\tau : \mathrm{simulate}(\sigma,\tau) \text{ reaches a SHIFT}\}$,
where \code{simulate} runs the reduce chain on a virtual overlay stack (pop
$|\mathrm{rhs}|$, GOTO, repeat) until shift or error. Cost is $O(|\mathrm{row}|
\times \mathrm{depth})$ worst case, amortized far less, and memoized per stack
node.

\paragraph{EOS gating by reduce-chain simulation, mid-lexeme aware.}
EOS may enter the mask only when the current output is a \emph{complete
sentence}. A raw row read is wrong twice: compressed tables have spurious
reduces, and after \code{...FROM t} the pending lexeme \code{t} is a complete
identifier still awaiting maximal-munch finalization---the stack alone would
wrongly report EOS illegal. The rule: EOS is legal iff (the remainder is empty
\emph{or} the pending lexeme finalizes as exactly one winning hypothesis) and,
after virtually shifting that finalized terminal on a scratch stack, ACCEPT is
reachable via the reduce chain of \code{\$end}.

\section{The token--terminal bridge}\label{sec:bridge}

LLM tokens do not align with grammar lexemes: \code{sel}+\code{ect} spells
\code{SELECT}, and schema identifiers are never single tokens. Masking out
``fragment'' tokens destroys completeness; admitting them requires scoring every
token's bytes against the live configuration. The early design snapshot
illustrated this misalignment and left the mechanism as later work; this section
is that mechanism.

\subsection{The trie walk and the exact mask}

\begin{definition}[Token trie]\label{def:trie}
The \emph{token trie} is a byte trie over $\{\mathrm{bytes}(t) : t \in V\}$,
stored as one DFS-contiguous array of packed 8-byte nodes (edge byte, token id,
subtree size) plus an alias table mapping each node to \emph{all} token ids
sharing its spelling. One trie per tokenizer fingerprint; special tokens are
excluded and permanently masked (EOS enters masks only via the explicit gate).
\end{definition}

\begin{definition}[The mask]\label{def:mask}
For configuration $\kappa$ with current output $w$,
\[
M(\kappa) \;=\; \{\, t \in V\setminus\{\mathrm{EOS}\} :
w\cdot\mathrm{bytes}(t) \in \Prefix(L) \,\}\;\cup\;
\{\, \mathrm{EOS} \ \text{iff}\ w \in L \,\}.
\]
\end{definition}

$M(\kappa)$ is computed by a DFS over the trie that carries an $O(1)$-updatable
scan state per frame (DFA state, current segment, last-accept, pending
forced-emission cascade): a byte that kills the DFA triggers maximal-munch
emission of the last accept and requeues the unconsumed tail. Rejection is
monotone under extension, so rejected subtrees are skipped wholesale via the
packed subtree size. At identifier positions the DFA runs against the L3 trie
intersection. Masks are complete over \emph{ids} via alias expansion.

\subsection{The context-independent/context-dependent split}
\label{sec:cicd}

\begin{proposition}[Boundary-crossing tokens are not $(\rho, A)$-determined]
\label{prop:cd}
Let token $t$'s bytes cross a terminal boundary \emph{and continue} (e.g.\
\code{'),'} or \code{'1;'}): they complete a terminal $\tau_1$ and begin
material beyond it. The viability of $t$ at $\kappa = (\sigma,\rho)$ depends on
the allowed-terminal set \emph{after shifting} $\tau_1$, i.e.\ on the stack
$\sigma$. Hence no cache entry keyed only on the lexer state and the current
allowed set $A$ can decide $t$ soundly.
\end{proposition}

The walk therefore classifies every admitted token as \textbf{CI}
(context-independent: resolvable within the current lexeme, or ending exactly at
one terminal boundary with that terminal in $A$)---cacheable---or \textbf{CD}
(context-dependent: boundary-crossing continuation)---returned as a residue list
and checked against the \emph{live stack} every step, never cached. Without this
split, the cache-key soundness obligation of \S\ref{sec:keys} is violated
\emph{by construction}, not merely unverified. The CD residue is the main
hot-path cost and gets its own machinery:

\paragraph{Verdict-equivalence grouping.}
The per-step verdict of a CD entry depends on the entry only through its
per-event candidate terminal sets, its segments and trailing remainder
\emph{only via the lexicon predicates}, and the live set of its remainder's scan
state. Entries are therefore partitioned once, at publish time, into
\emph{verdict-equivalence classes}; each step evaluates one stack-dependent
verdict per class and extends the mask with the class's alias-expanded ids.
On the SQL bench this turned $\sim$25k per-step entry checks into $\sim$150
class verdicts (measured mean 143--150 classes/step at $n{=}16{,}000$;
\S\ref{sec:eval-r}), and a later kernel round that keyed classes by verdict
equivalence instead of raw bytes collapsed $\sim$55k singleton groups to
$\sim$1.4k classes, a $9.3\times$ cold-build reduction (\S\ref{sec:kernels}).

\subsection{Configuration-keyed caching: OBL-KEY1 and the identifier rule}
\label{sec:keys}

The write-back mask cache carries the v0.0.5 table's spirit---precomputed where
cached, computed on miss---to its workable form: entries are keyed on
configurations, published idempotently (content-hashed
$\mathrm{entry\_id} = \mathrm{BLAKE2b\text{-}128}(\text{key} \,\|\, \text{encoding
tag} \,\|\, \text{payload})$, so racing writers produce the same id), immutable,
and versioned under namespaces that roll over when grammars are superseded.

\begin{definition}[T1 key]\label{def:key}
The per-grammar (T1) cache key of a configuration is
\[
k(\kappa) = \big(\,\mathrm{kind} \in \{\text{ident},\text{generic}\},\;
r,\; \mathrm{sorted}(A),\; \mathrm{schema\_fp} \big),
\]
where $r$ is the remainder bytes (which determine the lexer scan state), $A$ the
allowed-terminal signature from the live stack, and the schema fingerprint is
carried by identifier-position keys as a \emph{distinct key type}.
\end{definition}

\begin{proposition}[Cache-key soundness obligation, OBL-KEY1]\label{prop:oblkey}
Any two configurations sharing a key must produce byte-identical
context-independent masks: the key must refine the Myhill--Nerode classes of the
(lexer product-DFA $\times$ allowed-terminal set $\times$ identifier-lexicon)
product. The CD residue is exempt \emph{because it is never cached}.
\end{proposition}

OBL-KEY1 is verified, not assumed: differential tests check cache-on $\equiv$
cache-off over randomized replays including cross-role hits, and publish is
content-addressed so a racing writer of the same key with a different mask trips
a runtime assertion.

\begin{proposition}[Identifier composition rule]\label{prop:ident}
At identifier positions the mask must come from the L3 allow-list trie
intersection; unioning a cached generic-identifier verdict is unsound. A generic
verdict admits tokens spelling \emph{forbidden} identifiers, and the parser will
not reject them later---they are grammatical \emph{as identifiers}---so the
error would be a silent RBAC violation, invisible to any downstream syntactic
check.
\end{proposition}

The rule is enforced structurally: identifier-position keys are a distinct key
type that cannot collide with generic keys, and consulting a generic entry at an
identifier position raises an error in all builds (a test injects the condition
and asserts it fires).

\paragraph{The cross-family tier (T2) and the scanner normal form.}
A second tier shares entries across the grammar \emph{family}---the enterprise
shape is many roles $\times$ schemas over one dialect. Role projections share
the L1 terminal numbering (assigned at freeze; projections subset productions,
never renumber), so two roles reaching the same normalized configuration share
entries. To maximize sharing soundly, non-identifier keys normalize the
remainder to the \emph{genN scanner normal form}: scanning $r$ with last-accept
tracking yields $(q, \ell, p)$---the live DFA state $q$, the last-accept length
$\ell$, and the last-accept state $p$---and the key becomes
\[
\big(\text{genN},\; p,\; q,\; v,\; \mathrm{sorted}(A),\; \mathrm{schema\_fp}\big),
\qquad v = r[\ell{:}] \ \text{(the post-accept suffix bytes)},
\]
sound only under the \emph{lexicon-visibility guard}
$(\mathrm{live}[q] \cup \mathrm{accepts}[p]) \cap \mathrm{LEX} = \emptyset$:
every terminal whose lexicon predicates the walk could consult on
remainder-derived bytes lies in that set, so under the guard those checks are
lexicon-inert and the walk's future is a function of $(p,q,v,A)$ and the
lexicons alone---remainders the walk provably cannot distinguish share one
entry. The suffix $v$ is load-bearing (e.g.\ \code{b"1e"} and \code{b"1E"} share
$(q,\ell,p)$ but requeue different bytes on emission). Guard failure falls back
to the raw key.

\begin{remark}[A soundness catch, found by fuzzing the fix]\label{rem:t2}
When any lexicon exists, walk-time CD filtering embeds schema words into entry
\emph{content}, so the schema fingerprint must scope \emph{all} entries of a
lexicon-bearing producer, not only identifier positions. The original T2 tier
shared generic entries across schemas; a 50-seed shared-registry fuzz of the
genN work produced a counterexample (a whitespace remainder with
$A = \{\mathrm{LPAREN}\}$ served one schema's \code{(}-continuations of schema
words to another). The unsound legacy behavior survives only behind a kill
switch for replaying old audit logs. The differential fuzz ``failing'' against
the old behavior was the fuzz finally being given a correct oracle.
\end{remark}

\section{Guarantees}\label{sec:guarantees}

Each guarantee is stated with its exact preconditions; each clause is paired with
the test or benchmark that verifies it (\S\ref{sec:eval}). $L$ is the
role/schema-projected language of \S\ref{sec:prelim}.

\begin{proposition}[Soundness]\label{prop:sound}
Every emitted token keeps the detokenized output in $\Prefix(L)$.
\emph{Preconditions:} exact masks (no lookahead approximation); hard $-\infty$
masking (soft down-weighting leaves nonzero mass on illegal tokens and voids the
guarantee); the identifier composition rule (Proposition~\ref{prop:ident}).
\end{proposition}

\begin{proposition}[Completeness and dead-end freedom]\label{prop:complete}
No token is blocked whose byte string extends the current viable prefix toward a
member of $L$; consequently every viable prefix has at least one legal token and
generation cannot wedge. \emph{Preconditions:} (a) byte-complete vocabulary (all
byte values reachable via $\mathrm{bytes}$; verified per tokenizer adapter---degradation is an explicit warning that voids completeness, never soundness);
(b) reduced projected grammar; (c) exact trie walk with alias expansion (masks
over ids, not spellings); (d) validated lexicons $W_c \subseteq R_c$.
\end{proposition}

Precondition (d) earned its ``validated, not assumed'' status empirically: the
first real-world schema violation (a Spider database column named
\code{Official\_ratings\_(millions)}, whose parentheses lie outside the
column-name regex) produced an empty mask at a viable state within 100
generations---every prefix of the word passes the lexicon-prefix check, yet no
token can ever complete the lexeme. The precondition is now checked at guide
build.

\begin{proposition}[Termination]\label{prop:term}
In mode 1 (the \grid{}-owned decode loop): every non-error stop satisfies
$\mathrm{output} \in L$. \emph{Preconditions:} EOS-iff-ACCEPT gating
(\S\ref{sec:oracle}); a \emph{token-denominated} reserve---per-configuration
minimal-completion costs counted in vocabulary tokens (a terminal-denominated
reserve under-reserves: one identifier terminal can span many tokens), summed
incrementally on the stack---with the trigger
$\mathrm{budget} \le |\mathrm{completion}| + \mathrm{safety}$ answering
$\mathrm{Write}(\text{shortest legal completion} + \mathrm{EOS})$, never a bare
EOS away from ACCEPT; and a finite length budget.
In mode 2 (processor-only serving, where a logits processor cannot append
tokens), the guarantee \emph{weakens} to EOS-only-at-ACCEPT: reserve completion
is unavailable, truncation at budget exhaustion is possible, and the downgrade
is recorded in the audit seal.
\end{proposition}

\begin{proposition}[Complexity: requirement R]\label{prop:r}
Guard-rail cost is amortized $O(1)$ per token and $O(n)$ total; the per-step
worst case is bounded by \emph{nesting depth} (one terminal can trigger a reduce
cascade proportional to stack depth; SQL's prefix operators are inherently
right-recursive, so cascades cannot be linted away), never by the output
position $n$. Per-stream space is $O(\mathrm{depth})$ beyond the shared
artifacts. \emph{Preconditions:} nothing in the hot loop may read state
proportional to $n$---the processor keys per-row states incrementally (a
splitmix64 chain, not prefix hashing), the audit configuration hash is a rolling
$O(1)$-per-push mix carried in stack nodes, and mask cost is a function of the
configuration.
\end{proposition}

\begin{proposition}[The RBAC boundary]\label{prop:column-rbac}
Grammar masking enforces verb-level and table-level policy. Per-table
\emph{column} restrictions are not enforceable by any left-to-right CFG mask: in
SQL the SELECT list precedes FROM, so the alias$\to$table binding needed to
judge a column is unknown at column-mask time, and alias binding is
context-sensitive. Column-level policy is a post-parse semantic check on the
completed statement (one AST walk) or a view/rewrite layer in the database.
\end{proposition}

Proposition~\ref{prop:column-rbac} is not an apology but a load-bearing design
input: it defines the residue that the \code{SemanticChecker} names precisely,
which in turn is what makes checker-guided repair work at scale
(\S\ref{sec:eval-spider}).

\begin{remark}[Distribution faithfulness is not claimed]\label{rem:dist}
Per-step masking does not sample from $P(x \mid x \in L)$: it renormalizes
locally per step~\cite{park2024grammar}. \grid{} does not repair this by
default---the benchmark plan measures downstream execution accuracy instead of
pretending the gap away---and constraint-induced quality loss for small models
is real and documented~\cite{tam2024speak}. Grammar-aligned
decoding~\cite{park2024grammar} and two-phase reasoning-then-constrained
approaches~\cite{banerjee2025crane} are compatible add-ons, deferred by
decision.
\end{remark}

\section{System}\label{sec:system}

\subsection{Architecture}\label{sec:arch}

\begin{figure}[t]
\centering
\begin{tikzpicture}[
  font=\scriptsize,
  box/.style={draw, rounded corners=1pt, inner sep=3pt, align=center, fill=blue!4},
  art/.style={draw, inner sep=3pt, align=center, fill=orange!8},
  lab/.style={font=\scriptsize\bfseries, anchor=west},
  arr/.style={-{Stealth[length=2mm]}, semithick},
  node distance=4mm and 6mm
]
% ---- offline band ----
\node[lab] (offlab) at (-0.2,0) {OFFLINE / PER DEPLOYMENT};
\node[box, below=3mm of offlab.west, anchor=north west] (policy) {PolicyBundle\\(RBAC store)};
\node[box, right=of policy] (l2) {RoleProjection (L2)\\subset $\to$ reduce};
\node[box, below=of policy] (l1) {DialectGrammar (L1)};
\node[box, below=of l1] (schema) {SchemaSnapshot};
\node[box, right=of schema] (l3) {SchemaLexicon (L3)\\identifier tries};
\node[box, right=18mm of l2, yshift=-9mm] (compile) {compose $\to$ reduce\\$\to$ LALR(1) compile};
\node[art, right=of compile] (cg) {CompiledGrammar\\(fingerprinted)};
\node[box, below=9mm of schema] (tok) {TokenizerAdapter\\\code{token\_bytes}};
\node[art, right=of tok] (trie) {TokenTrie\\(per tokenizer fp)};
\node[art, right=of trie] (reserve) {ReserveTable\\(grammar fp, tokenizer fp)};
\draw[arr] (policy) -- (l2);
\draw[arr] (schema) -- (l3);
\draw[arr] (l2.east) -- (compile.west|-l2.east) -- (compile.north west);
\draw[arr] (l1.east) -- (compile.west);
\draw[arr] (l3.east) -- (compile.west|-l3.east) -- (compile.south west);
\draw[arr] (compile) -- (cg);
\draw[arr] (tok) -- (trie);
% ---- per request band ----
\node[lab, below=4mm of tok.south west, anchor=north west, xshift=-0mm] (reqlab) {PER REQUEST};
\node[box, below=3mm of reqlab.west, anchor=north west] (registry) {GrammarRegistry lookup\\(single-flight on miss, E17)};
\node[box, right=of registry] (guide) {GridGuide(compiled, trie,\\cache, audit)};
\node[box, right=of guide] (proc) {GridLogitsProcessor /\\vLLM structured-output backend};
\draw[arr] (registry) -- (guide);
\draw[arr] (guide) -- (proc);
% ---- per token band ----
\node[lab, below=4mm of registry.south west, anchor=north west] (toklab) {PER TOKEN (hot path)};
\node[box, below=3mm of toklab.west, anchor=north west] (stack) {parser stack $\sigma$\\allowed terminals $A$};
\node[box, right=of stack] (producer) {MaskProducer};
\node[box, right=of producer] (cache) {MaskCache T1/T2\\hit $\mid$ miss: trie walk\\+ write-back};
\node[box, right=of cache] (cd) {CD residue check\\(live stack, uncached)};
\node[box, below=of cd, xshift=-6mm] (sampler) {\code{masked\_fill\_($-\infty$)}\\$\to$ sampler};
\node[box, left=of sampler] (advance) {advance: lexer scan\\$\to$ shift/reduce};
\node[box, left=of advance] (audit) {AuditLog.append\\(config hash, entry id,\\token, blocked count)};
\draw[arr] (stack) -- (producer);
\draw[arr] (producer) -- (cache);
\draw[arr] (cache) -- (cd);
\draw[arr] (cd) -- (sampler);
\draw[arr] (sampler) -- (advance);
\draw[arr] (advance) -- (audit);
\draw[arr] (advance.south) -- ++(0,-4.5mm) -| ([xshift=-4mm]stack.west) -- (stack.west);
\end{tikzpicture}
\caption{\grid{} architecture at its three time scales (mirrors the normative
design document). Offline artifacts are immutable and fingerprinted; per-request
construction is single-flight; the per-token loop reads only
configuration-sized state.}
\label{fig:arch}
\end{figure}
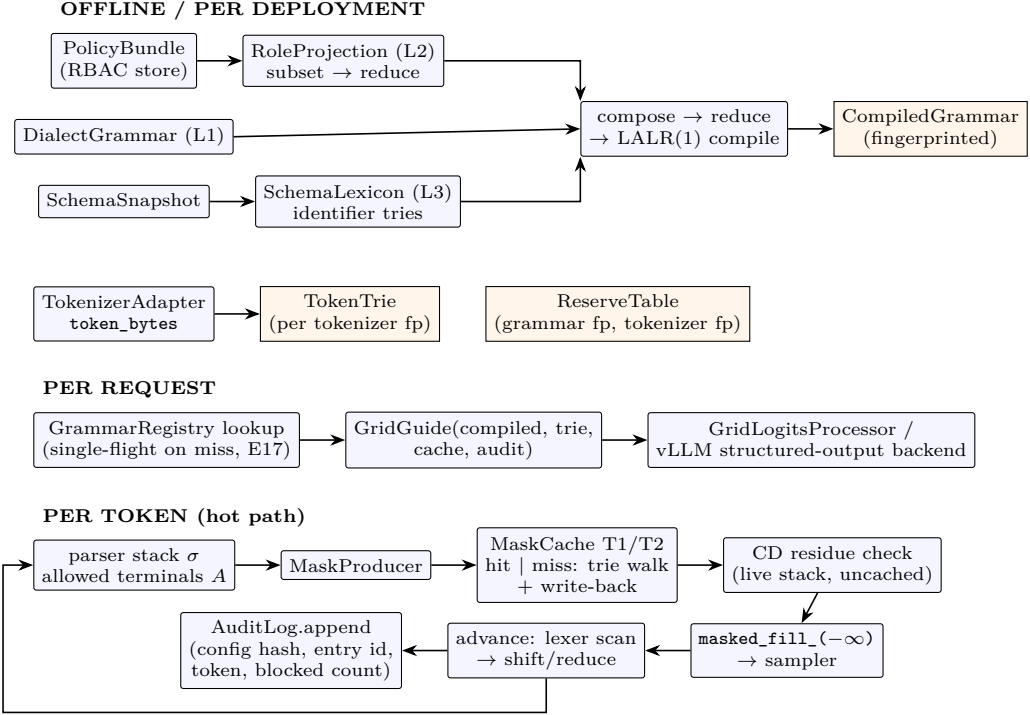

Figure~\ref{fig:arch} shows the three time scales. Everything expensive is
offline and content-addressed; everything per-request is a registry lookup with
single-flight construction; the per-token hot path touches only
configuration-sized state.

\paragraph{The per-token hot path} (normative order; every step audited):
\begin{enumerate}\itemsep1pt
\item $A \leftarrow$ allowed terminals by reduce-chain simulation on the virtual
stack (memoized per node).
\item $\mathrm{eos\_ok} \leftarrow$ mid-lexeme-aware ACCEPT-reachability
(\S\ref{sec:oracle}).
\item Reserve trigger (session budget, not grammar state): if
$\mathrm{budget} \le |\mathrm{completion}| + \mathrm{safety}$, return
$\mathrm{Write}(\text{shortest completion} + \mathrm{EOS})$.
\item Key $\leftarrow$ Definition~\ref{def:key} / the genN normal form; T1 then
T2 lookup; miss $\Rightarrow$ trie walk (CI mask $+$ CD residue), publish
idempotently.
\item Mask $\leftarrow$ CI $\cup$ CD-pass (live-stack class verdicts) $\cup$
$\{\mathrm{EOS}\ \text{iff eos\_ok}\}$; empty mask raises---a bug by
Proposition~\ref{prop:complete}, asserted throughout the test suite.
\item Singleton masks extend to a maximal forced span (jump-forward
\code{Write}, bounded chain length); otherwise return the \emph{full} exact
mask, applied as a hard in-place fill.
\item On the sampled token: bytes $\to$ lexer advance (maximal munch, candidate
sets, parser-viable pick) $\to$ shift/reduce on the persistent stack $\to$
append the audit record.
\end{enumerate}

Two implementations of this semantics coexist: a $\sim$140-line pure-Python
\emph{executable specification} (brute-force trial-parse over the vocabulary)
and the fast path, bound bit-identically by differential and parity tests. This
discipline caught---before any debugging session---alias ids silently dropped
from masks, lexicon-blind reserve completions, and a real Rust/Python divergence
in a CD group key.

\subsection{The kernel line: v4 $\to$ v5 $\to$ v5.1 $\to$ v6 $\to$ v7}\label{sec:kernels}

Four hot symbols (trie walk, CD verdicts, LALR advance, bitmask fill) are bound
to Rust kernels; the Python spec path remains and a flag forces it. The
performance history is a lesson in \emph{where} the cost actually was:

\begin{itemize}\itemsep1pt
\item \textbf{v4} --- a persistent, structurally interned stack arena with
cross-token memos and a one-call assembled hit pass: warm-hit p50
$12.9 \to 3.5$\us{} (dev host). The interning insight: once a hot path is in
Rust, the next win is usually \emph{not calling it}---parser configurations
recur massively across positions. v4 also releases the GIL on the cold walk;
measured main-thread liveness during a cold replay rose from $9\%$ (walk holding
the GIL: ``overlap'' was a lie) to $88\%$.
\item \textbf{v5} --- scheduler-side \code{fill\_bits}: the whole per-request
bitmask row (pre-packed CI bit words $+$ live CD bits $+$ EOS) written into
vLLM's buffer in one FFI call. Motivated by the first real batched run: batch-1
overhead $+1.26\%$ but batch-8 $+3984\%$ and batch-32 $+5151\%$---three stacked
serving-only defects (a fill path that skipped the warm kernel; a prefetcher
that scheduled \emph{every} successor onto one worker; per-request copies
re-registering every entry). Fixes in dependency order took the batch-32 step
from 708\ms{} to 16\ms.
\item \textbf{v5.1} --- verdict-equivalence CD grouping (Remark in
\S\ref{sec:cicd}): cold CD-heavy mask builds $124 \to 13.3$\ms{} ($9.3\times$);
warm fill p90 $38 \to 3.8$\us. Both prior theories of this cost
(``151k-vocab walks are superlinear,'' ``the warm gap is Python dispatch'')
failed under component-level measurement; the real causes were CD keys embedding
raw bytes and re-packing $\sim$47k CD ids per fill.
\item \textbf{v6} --- the whole per-request serving step in-kernel
(\code{session\_accept}/\code{session\_fill}, one FFI call each): warm serving
step $7.39 \to 1.33$\us{}/request (accept $6.16 \to 0.56$, fill
$1.22 \to 0.77$; local M-series measurement), restricted to audit-off serving paths;
audit-enabled and processor-mode guides keep v5. A 200-seed lockstep fuzz of v6
against v5 showed zero divergence and the pre-implementation review surfaced a
real v5 bug (post-COMPLETE state resurrection under speculative decoding), now
pinned by the differential suite.
\item \textbf{v7} --- the entire cold mask \emph{miss} moved into Rust
(\code{RustWalker.walk\_payload} $+$ \code{RustVerdicts.register\_blob}): walk,
group-blob, CI pack, adaptive encode, BLAKE2b \code{entry\_id}, and registration
in \emph{one} GIL-released call; Python receives a handle plus a thin
\code{MaskEntryV7} shell. The red-team's original premise---that the per-cold-entry
Python cost was \code{make\_entry}/encode/hash---was wrong on inspection (those
are microseconds); the real source of the measured 6.8--8.7\ms{} per boundary
entry was the \code{WalkResult} glue building \code{CDEntry} reps plus
$\sim$30--60k gc-tracked objects per miss, whose allocation triggered gen-2 GC
pauses \emph{inside} the walk. Keeping the miss in Rust removes both: the
per-boundary-entry cost falls $6.8$--$8.7 \to 0.003$\ms. \code{entry\_id} is
byte-identical (so the audit trail replays unchanged); \code{GRID\_V7} defaults on and \code{=0} is
byte-identical to v6. The win is localized to the serving cold-entry path---the
warm hit path, the R-microharness, and the engine-comparison latencies are v7-unchanged.
\end{itemize}

\subsection{The serving contract}\label{sec:serving}

Under vLLM, \grid{} implements the structured-output backend contract
(scheduler-side bitmask fill, rollback natural on persistent states) and a
logits-processor mode. The contract that keeps a batch healthy:

\begin{itemize}\itemsep1pt
\item \textbf{Overlap, cold-only.} Mask computation runs on CPU overlapped with
the GPU forward pass. The prefetcher schedules a successor's mask onto the
worker pool \emph{only when it is not already warm}: the warm steady state never
touches the pool (unconditionally scheduling every successor serialized the
steady state behind one worker's queue---one of the three batched-run defects).
\item \textbf{Skip-a-round defer.} If a request's mask is not ready at sampling
time (worst case: a cold identifier-position walk), that request is deferred
from the current scheduling round and rejoins the next step; co-batched requests
are never stalled, and an approximate mask is never substituted---by design there
is no over-approximating deadline fallback, because one over-admitted token
exits $\Prefix(L)$ and voids Proposition~\ref{prop:sound}. In vLLM this is
realized as a scheduler mask-readiness guard (one small patch site; upstream PR
drafted).
\item \textbf{Single-flight everything.} One build per artifact fingerprint;
$N$ concurrent waiters share the result or the same exception; failures are
negatively cached with a TTL.
\item \textbf{An honest negative result: warmup-at-admission fails on the GIL.}
The tempting alternative to the defer---warm the fresh request's masks while it
sits in the WAITING queue---was implemented and measured harmful: the tier work
is GIL-bound, so admission warmup starved the live engine (fresh-request TTFT
$10\times$ worse \emph{and} multi-second batch stalls). It defaults off. The
defer, plus genN key normalization and rayon-parallel cold walks (bit-identical;
$2.05\times$ at 8 threads locally), is the shipped cold-schema stack.
\end{itemize}

\subsection{The audit chain}\label{sec:audit}

Every step appends a record
$(\mathrm{step}, \mathrm{config\_hash}, \mathrm{mask\_entry\_id},
\mathrm{token}, \mathrm{blocked\_count}, \mathrm{kind}, \mathrm{prev\_hash})$---including each interior token of a forced span and the EOS tail record---chained by BLAKE2b from a genesis constant and sealed with the stop reason,
artifact fingerprints, and mode flags (e.g.\ processor-only downgrades). The
configuration hash is rolling and $O(1)$ per push,
\[
H(\mathrm{node}) = \mathrm{low}_{64}\,\mathrm{BLAKE2b\text{-}128}\big(H(\mathrm{parent})
\,\|\, \mathrm{u32}(\mathrm{state}) \,\|\, \mathrm{u32}(\mathrm{goto})\big),
\qquad H(\mathrm{root}) = 0,
\]
since hashing the stack from scratch
would be a hidden $\Theta(\mathrm{depth})$-per-token dependence. Because mask
entries are immutable and content-addressed and the hash pins the parser
trajectory, a log replays against archived artifacts to bit-identical masks;
\S\ref{sec:eval-audit} reports the full-scale replay measurement.

\section{Worked examples}\label{sec:examples}

\subsection{Masking mid-identifier under a schema lexicon}\label{sec:ex1}

\begin{figure}[t]
\centering
\begin{tikzpicture}[
  font=\scriptsize,
  n/.style={circle, draw, inner sep=1.2pt, minimum size=4mm},
  acc/.style={circle, draw, double, inner sep=1.2pt, minimum size=4mm},
  cur/.style={circle, draw, fill=blue!15, inner sep=1.2pt, minimum size=4mm},
  e/.style={-{Stealth[length=1.6mm]}, semithick},
  blk/.style={-{Stealth[length=1.6mm]}, semithick, red!70!black, dashed},
  lab/.style={font=\scriptsize}
]
\node[n] (root) at (0,0) {};
\node[n, right=7mm of root] (e1) {};
\node[n, right=7mm of e1] (m) {};
\node[n, right=7mm of m] (p) {};
\node[n, right=7mm of p] (l) {};
\node[cur, right=7mm of l] (o) {};
\node[n, right=7mm of o] (y) {};
\node[n, right=7mm of y] (y2) {};
\node[acc, right=7mm of y2] (s) {};
\node[acc, right=11mm of s] (pub) {};
\draw[e] (root) -- node[above, lab]{e} (e1);
\draw[e] (e1) -- node[above, lab]{m} (m);
\draw[e] (m) -- node[above, lab]{p} (p);
\draw[e] (p) -- node[above, lab]{l} (l);
\draw[e] (l) -- node[above, lab]{o} (o);
\draw[e] (o) -- node[above, lab]{y} (y);
\draw[e] (y) -- node[above, lab]{ee} (y2);
\draw[e] (y2) -- node[above, lab]{s} (s);
\draw[e] (s) -- node[above, lab]{\_public} (pub);
\node[n, below=6mm of e1] (orders) {};
\draw[e] (root) |- node[below, lab, pos=0.75]{orders} (orders);
\node[below=1mm of o, lab, blue!60!black] {scan state after \code{emplo}};
\node[above=5mm of s, lab] (t1) {\code{yees}: CI (boundary, $\mathrm{TN}\in A$)};
\draw[e, blue!60!black] (t1) -- (s);
\node[below=6mm of s, lab] (t2) {\code{yees where}: CD (crosses boundary, live-stack verdict)};
\draw[e, blue!60!black] (t2.north) to[out=90,in=-45] (s.south east);
\node[below=1.5mm of t2, lab, red!70!black] (t3) {\code{yer}: blocked (no allowed word; whole subtree pruned)};
\draw[blk] (t3.west) to[out=170,in=-70] (y.south);
\end{tikzpicture}
\caption{Example~\ref{sec:ex1}: the L3 lexicon trie for
$W_{\mathrm{TABLE}}$ during the walk at remainder \code{emplo}. Double circles
are allowed-word accepts. Candidate tokens are classified CI
(cacheable), CD (checked against the live stack, never cached), or blocked
(whole trie subtrees pruned monotonically).}
\label{fig:ex1}
\end{figure}
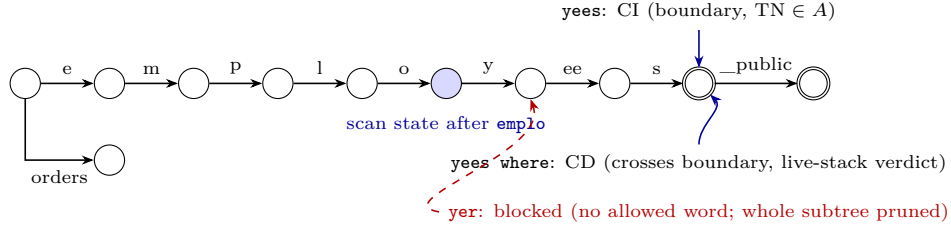

Take the CRUD-subset grammar (lowercase keywords; \code{TABLE\_NAME} and
\code{COLUMN\_NAME} share the regex \code{[a-z\_][a-z0-9\_]*} and are L3
categories), the role \code{analyst} whose L2 projection keeps only
\code{select\_stmt} among the query productions, and the schema lexicon
\[
W_{\mathrm{TABLE}} = \{\code{employees},\ \code{employees\_public},\
\code{orders}\},
\]
with a table \code{salaries} existing in the database but absent from the
role's lexicon. The model has emitted
\begin{center}
\code{select * from emplo}
\end{center}
so the configuration $\kappa$ has: stack $\sigma$ with \code{select}, \code{*},
\code{from} shifted (allowed next terminal set $A = \{\mathrm{TABLE\_NAME}\}$),
and remainder $r = \code{emplo}$ scanning inside an identifier lexeme with the
L3 category context set. Because $A$ contains an identifier category, the cache
key is the \emph{ident} type and carries the schema fingerprint
(Proposition~\ref{prop:ident}); the walk intersects the token trie with the
lexicon trie from the state reached by \code{emplo} (Figure~\ref{fig:ex1}).
Representative verdicts:

\begin{itemize}\itemsep1pt
\item token \code{yees} (bytes \code{yees}): completes exactly
\code{employees}, a lexeme boundary with $\mathrm{TABLE\_NAME} \in A$
$\Rightarrow$ \textbf{CI, admitted} (and cached in the entry's CI payload).
\item token \code{y}: extends to \code{employ}, a live lexicon prefix
$\Rightarrow$ \textbf{CI, admitted} (mid-lexeme continuation).
\item token \code{yees where}: crosses the boundary (emits
\code{TABLE\_NAME = employees}, skips ignored whitespace, begins
\code{where}) and continues $\Rightarrow$ \textbf{CD}: its verdict depends on
whether \code{where} can shift \emph{after} \code{table\_ref}, i.e.\ on the
live stack---checked this step, never cached (Proposition~\ref{prop:cd}). Here
it passes.
\item token \code{yer}: \code{employer} is not a prefix of any allowed word
$\Rightarrow$ \textbf{blocked}, monotonically with its whole trie subtree.
\item At the earlier position \code{select * from } (empty remainder): token
\code{sal} matches the generic identifier regex but is a prefix only of the
absent \code{salaries} $\Rightarrow$ \textbf{blocked}. A generic-IDENT cache
entry would have admitted it and the parser would never have objected---the
silent RBAC violation the identifier composition rule exists to prevent.
\item At the statement head, tokens spelling \code{insert}, \code{update},
\code{delete} are blocked for this role because L2 removed those productions
and reduction pruned their terminals: the verbs are not merely improbable,
they are \emph{outside the language}. No prompt can path to them (the
model-independent RBAC test probes every reachable identifier position with an
exhaustive multi-token speller; \S\ref{sec:eval-rbac}).
\item EOS is \textbf{not} in the mask: simulating \code{\$end} after virtually
finalizing \code{emplo} as \code{TABLE\_NAME} does not reach ACCEPT (the
grammar requires the closing \code{;}), even though \code{emplo}'s
finalization itself is legal. After a later \code{... ;} the reduce chain
reaches ACCEPT and EOS enters the mask through the explicit gate only.
\end{itemize}

\subsection{A cold schema entering a hot batch}\label{sec:ex2}

Thirty-two requests are decoding at steady state (warm caches, kernel fills of a
few microseconds) when a request arrives for a schema the deployment has never
seen. The fresh request pays, in order: a single-flight grammar+lexicon
specialization (27.3\ms{} cold, measured; every concurrent duplicate waits
on the same build), then---at its first identifier-heavy positions---cold trie
walks that are milliseconds-scale by nature (vocabulary-sized DFS). The contract
of \S\ref{sec:serving} makes these costs \emph{private to the fresh request}:
the walk runs GIL-released on the pool, overlapped with the forward pass; if it
is still unfinished at sampling time, the scheduler's mask-readiness guard
defers only that request for the round (Figure~\ref{fig:defer}). Measured on the
declared H100 SXM5 runner (\S\ref{sec:eval-serving}): at kernel v7 the serving
record holds the 31 warm co-tenants to a 15.3\ms{} worst engine step while the
fresh request's $\sim$0.66\,s cold window induces a transient $+33.8\%$ co-batched
TPOT slowdown---no longer a GIL-bound software cost (the v7 fused cold-miss path
removed that) but host CPU/memory-bandwidth contention between the cold walk and
the forward loop, which shrinks as walk threads grow. The defer remains the
load-bearing protection: with it disabled
the same leg reads $+373\%$ and 65.7\ms. The fresh request itself sees TTFT
$0.7$\ms{} and then runs at $1.00\times$ warm speed: a never-seen schema trades
its own first-token latency for exact masks throughout---no approximation is
ever substituted.

\begin{figure}[t]
\centering
\begin{tikzpicture}[font=\scriptsize, xscale=0.98]
% fresh window: the only span where co-tenants see contention (background band)
\fill[red!8] (2.6,0.08) rectangle (5.4,2.7);
\draw[red!60!black, |-|] (2.6,-0.62) -- (5.4,-0.62)
  node[midway, below=0.5mm] {\scriptsize fresh window ($\sim$0.66\,s)};
% axis
\draw[-{Stealth[length=2mm]}] (0,0) -- (13.3,0) node[below left, yshift=-1mm]{engine steps};
\foreach \x/\l in {1/k, 3/k+1, 5/k+2, 7/k+3, 9/k+4, 11/k+5}
  \draw (\x,0) -- (\x,-0.08) node[below]{$\l$};
% warm lane
\node[anchor=east] at (-0.1,1.1) {31 warm requests};
\foreach \x in {1,3,5,7,9,11}
  \fill[blue!25] (\x-0.35,0.85) rectangle (\x+0.35,1.35);
\node[anchor=east] at (13.2,0.6) {\scriptsize warm fills stay \us-scale; contention bounded to the fresh window (\S\ref{sec:eval-serving})};
% fresh lane
\node[anchor=east] at (-0.1,2.3) {fresh schema req.};
% admission
\fill[orange!35] (0.2,2.05) rectangle (1.6,2.55);
\node[anchor=west] at (0.1,3.25) {\scriptsize single-flight specialize (27.3\ms)};
\draw[gray] (0.9,3.05) -- (0.9,2.6);
% cold walk spanning step k+1..k+2
\fill[red!25] (2.6,2.05) rectangle (5.4,2.55);
\node at (4.0,2.9) {\scriptsize cold ident-position walk (on pool, GIL-released)};
% defer note, tied to the walk box above it (rounds k+1, k+2)
\node[red!70!black] at (4.0,1.68) {\scriptsize mask not ready $\Rightarrow$ deferred this round};
% rejoin at k+3
\fill[blue!25] (7-0.35,2.05) rectangle (7+0.35,2.55);
\fill[blue!25] (9-0.35,2.05) rectangle (9+0.35,2.55);
\fill[blue!25] (11-0.35,2.05) rectangle (11+0.35,2.55);
\draw[-{Stealth[length=1.6mm]}, blue!60!black, thick] (5.4,2.3) to[out=0,in=180] (6.65,2.3);
\node[anchor=west] at (6.7,3.25) {\scriptsize rejoins next round; then $1.00\times$ warm (TTFT $0.7$\ms)};
\draw[gray] (8.6,3.05) -- (7.15,2.62);
\end{tikzpicture}
\caption{Example~\ref{sec:ex2}: the cold-schema-into-hot-batch timeline. The
fresh request's compile and cold walks run off the critical path; the
mask-readiness guard defers only the fresh request from the rounds where its
mask is late ($k{+}1$, $k{+}2$). Co-batched requests never wait, and no
approximate mask is ever substituted; the shaded band is the fresh window,
the only span in which they see (bounded) contention.}
\label{fig:defer}
\end{figure}

\section{Evaluation}\label{sec:eval}

All numbers below are from committed benchmark reports in the repository
(\code{bench/RESULTS*.md}) or the measured-results log; each table carries its
host label verbatim. \emph{Declared runner} = a named cloud instance type and
image recorded in the report (Lambda 1$\times$H100 PCIe or SXM5 80GB, or
1$\times$A10 24GB; Ubuntu 24.04; virtualized); \emph{local dev} = an unpinned
Apple-Silicon workstation. Cross-engine \emph{ratios} proved host-invariant;
absolute constants carry the host label.

\subsection{Per-token mask latency vs.\ XGrammar, llguidance, Outlines}
\label{sec:eval-engines}

The engine-comparison harness measures the wall time to produce the full
next-token mask at each replay step, for all engines on the same SQL-subset
grammar (expressed in each engine's native format) and tokenizer, 11 replays.
Table~\ref{tab:engines} reports both tokenizers on the declared H100 SXM5
runner at kernel v7 (re-recorded; within noise of v6, as expected---v7's win is
localized to the serving cold-entry path, not this per-token latency path).

\begin{table}[t]
\centering\small
\caption{Per-token mask latency, SQL-subset grammar, kernel v7. Host: Lambda
1$\times$H100 SXM5 80GB, Ubuntu 24.04 (declared runner). Top: \code{gpt2}
(491 steps); bottom: Qwen2.5-0.5B-Instruct tokenizer (151k vocab; 509 steps).
``Rejected'' counts language-parity corners between grammar encodings
(maximal munch vs.\ explicit whitespace), not correctness bugs.}
\label{tab:engines}
\setlength{\tabcolsep}{4pt}\footnotesize
\begin{tabular}{@{}lrrrrrr@{}}
\toprule
engine & compile & p50 & p90 & p99 & slope (\us/pos) & rej.\\
\midrule
\multicolumn{7}{@{}l}{\emph{gpt2}}\\
\grid{} (Rust kernels: walk + CD + LALR) & 378.0\ms & 3.6\us & 80.0\us & 5{,}347.4\us & $-9.292$ & 0\\
XGrammar 0.2.3 (EBNF) & 94.1\ms & 72.2\us & 7{,}503.3\us & 25{,}586.7\us & $-43.759$ & 0\\
llguidance 1.7.6 (lark, driven directly) & 285.9\ms & 6.6\us & 223.9\us & 351.7\us & $-1.176$ & 2\\
Outlines 1.3.1 (CFG backend = llguidance) & 22{,}115.1\ms & 73.9\us & 431.5\us & 582.4\us & $-2.152$ & 2\\
\midrule
\multicolumn{7}{@{}l}{\emph{Qwen2.5-0.5B-Instruct (151k vocab)}}\\
\grid{} & 1{,}297.2\ms & 6.7\us & 109.0\us & 15{,}906.2\us & $-48.903$ & 0\\
XGrammar 0.2.3 & 342.7\ms & 588.6\us & 10{,}026.9\us & 31{,}774.6\us & $-63.015$ & 0\\
llguidance 1.7.6 & 979.5\ms & 14.9\us & 384.4\us & 1{,}200.2\us & $-1.819$ & 1\\
Outlines 1.3.1 & 13{,}796.8\ms & 61.7\us & 459.5\us & 558.7\us & $-2.215$ & 1\\
\bottomrule
\end{tabular}
\end{table}

Reading the table honestly: at kernel v7 \grid{} leads llguidance---the
strongest prior engine in our measurements---at \emph{both} p50 and p90 on
both tokenizers ($3.6$ vs.\ $6.6$\us{} p50 and $80$ vs.\ $224$\us{} p90 on
gpt2; $6.7$ vs.\ $14.9$\us{} and $109$ vs.\ $384$\us{} on Qwen), beats
XGrammar at every reported percentile, and is the only engine with zero
rejected replays. Its cache split is p50 $3.5$\us{} hit / $4.8$\ms{} miss at
$92\%$ hit rate (gpt2; $6.7$\us{} / $13.9$\ms{} / $92\%$ on Qwen), so the p99
remains cold-miss trie walk on this recursive SQL grammar: llguidance's
Earley/derivative core keeps the flattest p99 ($352$\us{} vs.\ \grid{}'s
$5.3$\ms{} on gpt2), and we say so plainly. Outlines $\ge$1.x has no CFG
engine of its own: \code{outlines.types.CFG} routes to a backend, default
llguidance, so its row is the same matcher plus Outlines' logits-processor
wrapper---identical rejects by construction. The steep negative table slopes
are an artifact of cold misses clustering early in replays; the warm-replay
check (slope $+0.002$\us/pos on gpt2, first-half p50 3\us{} vs.\ second-half
4\us; $-0.004$\us/pos and 5\us{} vs.\ 5\us{} on Qwen) is the R-relevant
statistic.

\subsection{Requirement R: flat per-token cost}\label{sec:eval-r}

\paragraph{The R microharness} (declared H100 SXM5 runner, kernel v7;
\code{gpt2}; $n = 16{,}000$ tokens per stream, 20 seeded runs per nesting depth,
warm-pass OLS; $\varepsilon = 10^{-4}$\us/pos) isolates guard cost from the model
(Table~\ref{tab:r}). Slope confidence intervals sit well over an order of
magnitude under $\varepsilon$ at every depth; hit p50 is $4.8$--$6.8$\us{}; the
steady-state hit rate is $100\%$; cumulative-cost fits are linear with
$R^2 \ge 0.9998$.

\begin{table}[t]
\centering\small
\caption{R microharness: mask latency vs.\ position, no model. Host: Lambda
1$\times$H100 SXM5 80GB, Ubuntu 24.04 (declared runner), kernel v7;
\code{gpt2}; $n{=}16{,}000$/stream; 20 seeded runs per depth (320{,}000 steps
each). GC is disabled during the warm timed pass.}
\label{tab:r}
\setlength{\tabcolsep}{4pt}\footnotesize
\begin{tabular}{@{}rlrrrrrr@{}}
\toprule
depth & slope (\us/pos, $\pm$95\% CI) & warm p50 & hit p50 & miss p99 & hit rate & cum.\ $R^2$ & CD cls./step\\
\midrule
0 & $-0.000004 \pm 0.000002$ & 3.7\us & 4.8\us & 6.56\ms & 100.0\% & 0.99985 & 143\\
4 & $-0.000007 \pm 0.000002$ & 4.3\us & 5.6\us & 5.84\ms & 100.0\% & 0.99992 & 148\\
8 & $-0.000007 \pm 0.000002$ & 4.7\us & 6.1\us & 6.11\ms & 100.0\% & 0.99995 & 149\\
16 & $-0.000006 \pm 0.000002$ & 6.0\us & 6.8\us & 6.31\ms & 100.0\% & 0.99994 & 150\\
\bottomrule
\end{tabular}
\end{table}

\paragraph{Head-to-head scaling vs.\ \code{guidance}, three eras} (local dev
Mac, unpinned; slopes and shape are the claim, absolutes indicative). The
project's founding claim---guard cost flat as generated context grows---was
measured against \code{guidance} in three vintages on the same growing
WHERE-chain statements, $n \in \{512, \ldots, 16{,}384\}$. At $n = 16{,}384$:
\grid{} runs at $3.2$\us/step with slope $-3.3\times10^{-5}$\us/pos (kernels
active); guidance 0.3.1 (today's llguidance core) is also flat but at a
$\sim$30$\times$ higher constant ($97.7$\us); guidance 0.1.5 (November 2023,
Python Earley) holds a flat \emph{median} ($\sim$114\us) while its worst-case
single step grows $67.5 \to 106.2$\ms{} across one generation's quarters
(gen-2 GC scanning the growing Earley chart, verified via \code{gc.callbacks});
and guidance 0.0.64 (July 2023, the release the v0.0.5 design was conceived
against) \textbf{breaks requirement R outright}: measured overhead-vs-position
slopes $+1{,}105$ to $+1{,}189$\us/pos (mechanism confirmed in its code:
full-string regex rebuilds per candidate per token, whole-prompt re-encoding per
operation), i.e.\ quadratic total cost---it spent 898\,s reaching position 873
and could not complete $n = 2{,}048$; the linear fit extrapolates (labeled as
such, not measured) to $\approx$19.5\,s/token at position 16{,}384. Total
constraint cost over one 16k replay: \grid{} \textbf{0.05\,s} vs.\ 1.88\,s
(0.3.1) vs.\ 2.52\,s (0.1.5) vs.\ $\approx$44\,h (0.0.64, extrapolated).

\subsection{MaskBench (JSON Schema)}\label{sec:eval-maskbench}

\grid{} claims extensibility to any LALR(1)-parsable language; MaskBench
(guidance-ai/jsonschemabench~\cite{geng2025jsonschemabench}) is the cheapest
public test of where that is true. \grid{} enters via a
JSON-Schema$\to$grammar compiler and a protocol-exact runner (TTFM/TBM semantics
verbatim; llama-3.1 tokenizer; 315-schema stratified sample; host: local dev,
unpinned). Table~\ref{tab:maskbench} shows the three-way comparison.

\begin{table}[t]
\centering\small
\caption{MaskBench, 315-schema stratified sample, llama-3.1 tokenizer, local
dev host (unpinned). TBM = time between masks; TTFM = time to first mask
(compile). Compile errors are \emph{declared} non-support; validation errors
(valid instance rejected) and invalidation errors (invalid accepted) are silent
correctness gaps.}
\label{tab:maskbench}
\begin{tabular}{@{}lrrr@{}}
\toprule
metric & \grid{} & llguidance & XGrammar (compliant)\\
\midrule
TBM avg & 562\us & 20\us & 101\us\\
TBM p50 & 28\us & 10\us & 10\us\\
TBM p75 & 34\us & 21\us & 29\us\\
TBM p90 & 75\us & 31\us & 57\us\\
TBM p99 & 7.7\ms & 179\us & 2.7\ms\\
TBM max & 11.7\ms & 2.0\ms & 12.0\ms\\
TTFM p50 & 6.0\ms & 0.31\ms & 2.4\ms\\
TTFM p75 & 8.1\ms & 0.44\ms & 209\ms\\
TTFM p99 & 302\ms & 7.7\ms & 13.0\,s\\
\midrule
compile errors (declared) & 79 & 62 & 0\\
validation errors (silent) & \textbf{0} & 3 & 27\\
invalidation errors & 68 & 0 & 37\\
\bottomrule
\end{tabular}
\end{table}

\grid{}'s p25--p75 (14/28/34\us) is the kernel hit path, with kernels active on
$100\%$ of compiled schemas, and \grid{} is the only engine with \emph{zero}
validation errors---every valid instance of every schema it compiled was
accepted; its 68 invalidation errors are all traceable to deliberately ignored
value constraints (the XGrammar-default convention, itemized in the report).
Its honesty boundary is llguidance-style: 79/315 schemas are declared compile
errors (allOf, patternProperties, if/then/else, \dots, including 5 genuine LALR
conflicts). The TBM p90 traces a clean kernel lineage across the identical
315-schema sample: v3-era $27.8$\ms{} $\to$ v5.1 verdict-equivalence grouping
$208$\us{} $\to$ v7 fused walk$\to$blob$\to$register $75$\us. This is the
kernel-v7 re-record: the correctness columns are byte-identical to the earlier
records, while the last step---v5.1 to v7---cut TBM p90 a further $\sim$$2.8\times$
($208 \to 75$\us) and the TBM average from $683$ to $562$\us{} by moving the
per-cold-entry Python materialization/glue (and its gen-2 GC pauses) into Rust;
the llguidance and XGrammar latency arms moved under $10\%$ between records, a
host-noise control placing the delta on \grid{}'s side. Two structural facts
explain the remaining tails: the TBM p95$+$ tail is the cold trie walk, which
MaskBench's one-shot-per-schema protocol never lets the write-back cache amortize
(the serving benchmark is where that design choice pays), and TTFM is a pure-Python
table build ($19\times$ llguidance's at p50, but 26--43$\times$ ahead of
XGrammar's p75$+$ compile blowups).

\subsection{Spider: execution accuracy and checker-guided repair}
\label{sec:eval-spider}

Executability is measured where it matters: the full 1{,}034-question Spider
dev set~\cite{yu2018spider}, greedy decoding, grammar with $100\%$ dev-gold
coverage plus per-database L3 lexicons---every constrained output parses with
schema-valid identifiers by construction (Table~\ref{tab:spider}).

\begin{table}[t]
\centering\small
\caption{Spider dev execution accuracy. 7B rows: Qwen2.5-7B-Instruct, full
1{,}034 questions (\code{grid}/unconstrained on the declared H100 PCIe runner;
the repair pair reproduced \code{grid} exactly on the declared A10 runner). 0.5B
rows: Qwen2.5-0.5B-Instruct, 100 questions, local dev host. EX = result-set
match on the Spider databases.}
\label{tab:spider}
\begin{tabular}{@{}llrrrrr@{}}
\toprule
model & arm & executes & EX & EX delta & truncated & tok/query\\
\midrule
0.5B ($n{=}100$) & \grid{} (repair inert at this scale) & 57.0\% & 29.0\% & $+13.0$ & 4.0\% & 41\\
0.5B & unconstrained & 31.0\% & 16.0\% & --- & 9.0\% & 61\\
\midrule
7B ($n{=}1034$) & \textbf{\grid{} (with repair)} & \textbf{94.5\%} & \textbf{55.2\%} & $\mathbf{+2.3}$ & 0.5\% & 40\\
7B & \grid{} (repair off, ablation) & 91.3\% & 53.7\% & $+0.8$ & 0.9\% & 35\\
7B & unconstrained & 91.0\% & 52.9\% & --- & 0.2\% & 33\\
\bottomrule
\end{tabular}
\end{table}

The scale finding, measured at both ends: the mask alone is worth $+13$ EX
points at 0.5B---it erases the syntax-error class a weak model commits
constantly ($+26$ points of syntactic validity)---but only $\sim$$+1$ at 7B,
because a capable model rarely errs syntactically. The repair half closes the
loop at scale: \grid{}'s residual 7B failures are alias$\leftrightarrow$column
binding---exactly the class Proposition~\ref{prop:column-rbac} proves out of
mask scope, and exactly what the alias-aware \code{SemanticChecker} names
precisely. One constrained retry with the violations quoted back converts a
third of that floor ($91.3 \to 94.5\%$ executes; $21\%$ of kept retries newly
correct) at $+5$ tokens/query ($+14\%$) on the $\sim$7\% of queries that
engage. The \emph{capability symmetry} is measured in both directions, not
asserted: at 0.5B the same feedback is worthless (an interim run of 55
questions: repair metrics identical to plain \grid{}---the model repairs its
binding mistakes into different-but-equally-wrong queries), while at 7B the
mask matters less but the feedback converts---constraint quality determines
feedback quality, because the checker can only name violations precisely when
the mask has already guaranteed everything else. Ablations on the same harness
(EX-invariant by construction): disabling the write-back cache costs $32\%$ of
generation throughput ($2.5 \to 1.7$ tok/s, $n{=}20$); audit-off and
jump-forward-off move throughput within noise.

\subsection{Batched serving under vLLM}\label{sec:eval-serving}

The end-to-end serving benchmark runs vLLM~0.24 (V1 engine) with Qwen2.5-7B on the
declared H100 SXM5 runner: heterogeneous grammars (4 distinct) across batches
1/8/32, unconstrained control arm, warm-through protocol, kernel v7
(Table~\ref{tab:serving}).

\begin{table}[t]
\centering\small
\caption{Serving record, kernel v7. Host: Lambda 1$\times$H100 SXM5 80GB,
Ubuntu 24.04 (declared runner), vLLM 0.24, Qwen2.5-7B, heterogeneous grammars.
The cold-miss stress arm injects a fresh never-warmed schema into a warm
batch-32; its metric is per-request TPOT over the 31 warm co-batched requests,
computed with artifact-robust estimators (median-over-legs degradation;
min-over-legs max step) for the reason given in \S\ref{sec:eval-serving}.}
\label{tab:serving}
\begin{tabular}{@{}lr@{}}
\toprule
TPOT overhead vs.\ unconstrained, batch 1 & $+0.15\%$\\
TPOT overhead, batch 8 & $+0.73\%$\\
TPOT overhead, batch 32 & $\mathbf{+1.51\%}$\\
TTFT, cold role+schema specialize & 27.3\ms\\
TTFT, warm & 1.51\ms\\
\midrule
\multicolumn{2}{@{}l}{\emph{Cold-miss stress arm (fresh schema into warm batch-32):}}\\
max engine-step wall, min over legs & $\mathbf{15.3}$\ms\\
\quad raw per-leg maxima & $[15, 15, 16, 16, 16]$\ms\\
co-batched TPOT slowdown during cold window & $\mathbf{+33.8\%}$\\
fresh request: TTFT / completion / eff.\ TPOT & 0.7\ms{} / 663.0\ms{} / 6.97\ms{} ($1.00\times$ warm)\\
\midrule
concurrent cold start (single-flight) & 1 build / 8 waiters; same error on failure\\
\bottomrule
\end{tabular}
\end{table}

Warm serving overhead is low at every batch size---$+1.51\%$ time-per-output-token
at batch~32---warm and cold TTFT are $1.51$ and $27.3$\ms, and both single-flight
behaviors hold. In the cold-miss stress arm the warm co-tenants' worst engine step
is 15.3\ms{} (thread-invariant at 15--18\ms{} across walk threads $0/2/8$). The one
residual cost is a transient co-batched TPOT slowdown of $+33.8\%$, confined to the
fresh schema's cold window; we characterize it below.

What v7 changed. On the v6 stack the cold window cost more---a $+114.7\%$
co-batched slowdown and a 36.0\ms{} worst step---and both were attributed to
GIL-bound mask-entry materialization in the fresh request's cold window. That
attribution was half right. Moving the entire cold miss into Rust
(\S\ref{sec:kernels}: fused walk$\to$blob$\to$register, one GIL-released call)
collapsed the per-boundary-entry cost $6.8$--$8.7 \to 0.003$\ms{} and, with it,
the worst engine step on real hardware ($36.0 \to 15.3$\ms). The residual glue theory
(``\code{make\_entry}/encode/hash is the cost'') was disproved by the fix: the
real culprit was the \code{WalkResult} glue allocating $\sim$30--60k gc-tracked
objects per miss, whose gen-2 pauses landed \emph{inside} the walk---which is
exactly why keeping the walk in Rust removed them.

The estimators. The cold-window metrics are computed with artifact-robust
estimators (median-over-legs degradation; min-over-legs max step, with the raw
per-leg maxima always printed) because vLLM~0.24's multiprocess engine exhibits a
once-per-leg 0.7--2\,s frozen engine step that we exonerated five ways before
believing it: it fires in all-warm baselines with zero \grid{} work in flight,
is invariant to the defer, warmup, walk-thread, and JIT-warming levers and to
both child- and driver-side GC control, and vanishes entirely with the engine
in-process---an engine-topology artifact, filed upstream as vllm-project/vllm
issue \#48229 (the repository's LESSONS.md~\S6.8--6.9 holds the exoneration
record). The raw per-leg step maxima $[15, 15, 16, 16, 16]$\ms{} show the v7
max step is now clean and tight, with no exogenous 1--2\,s outliers polluting
the min-over-legs estimator.

The cold-window slowdown, characterized. At
v6 the co-batched degradation was called a Python/GIL/software cost with a
kernel fix pending. That fix has shipped, and the degradation did not go to
zero---because it was never primarily a software cost. It is genuine host
CPU/memory-bandwidth contention between the cold walk and the engine forward
loop during the fresh schema's $\sim$0.66\,s specialization window. The
diagnostic evidence is direct: the slowdown \emph{decreases} as walk threads
increase, because more threads shrink the window in which the contention
occurs---a software-serialization cost would not behave this way. Walk/pool
thread niceness mitigates the residual. Full closure is therefore a
compute-isolation trade-off (throttle the walk and lengthen the fresh request's
TTFT, or accept the number), noted as future work rather than a defect to fix.
Throughout, the fresh request itself pays only its own cost: TTFT $0.7$\ms, and
$1.00\times$ warm effective TPOT once specialized---there is zero steady-state
co-tenant interference. As a pre-v7 observation (superseded by this
record, retained only for lineage): an artifact-free JIT-warmed lockstep
leg on the v6 stack read $-1.75\%$ co-batched slowdown and a 23.9\ms{} max
step; the same leg with the defer disabled read $+373\%$ and 65.7\ms, the
attribution that makes the defer the load-bearing protection. Finally, the
tempting admission-time warmup alternative was measured harmful on the GIL and
ships disabled (\S\ref{sec:serving}); negative results are recorded next to the
positive ones.

\subsection{Policy enforcement and end-to-end correctness at scale}
\label{sec:eval-rbac}

The guarantees of \S\ref{sec:guarantees} are checked directly, each against the
test that exercises the corresponding property.

\paragraph{Soundness, completeness, termination end-to-end.} Two arms exercise
the whole decode loop. The walk arm generates 10{,}000/10{,}000 outputs that
parse, with 0 dead ends and 0 budget overruns (local dev). The model arm, driving
Qwen2.5-0.5B on the declared H100 runner, generates 1{,}000/1{,}000 parseable and
audit-verified outputs, again with 0 dead ends, and 706 reserve-completed stops
(the termination machinery of Proposition~\ref{prop:term} firing as designed).

\paragraph{Policy/RBAC enforcement.} A model-free exhaustive speller probes every
reachable identifier position: across 58 role$\times$position$\times$target probes
it records \textbf{0 bypasses}, with 9/9 positive controls reachable so the test
is non-vacuous. A 12-prompt injection suite (the adversarial-prompt arm) emits 0
forbidden lexemes on the declared H100 runner, and column-violation fixtures are
100\% flagged by the post-parse checker (the residue
Proposition~\ref{prop:column-rbac} places out of mask scope).

\paragraph{Differential correctness, cache soundness, and kernel parity.} The
fast path is checked bit-exact against the trial-parse oracle over quota-counted
corpora including multi-terminal tokens, across a four-tokenizer matrix; the
viable-prefix oracle is checked by corpus, mutation, and mid-lexeme EOS
differentials against lark. Cache-key soundness (OBL-KEY1) is verified by
cache-on $\equiv$ cache-off runs including cross-role hits, namespace rollover,
and racing publishes. Grammar-pipeline reducedness and deterministic fingerprints
have property tests. All of these run in CI.

\subsection{Audit replay and tamper detection}\label{sec:eval-audit}

Over 1{,}000 generations spanning 29{,}242 audited steps and one cache-namespace
rollover, every log replays bit-identically (1{,}000/1{,}000) against its archived
artifacts, and tamper detection catches every injected mutation
(1{,}000/1{,}000), on local dev.

\subsection{Cross-engine comparison, summarized}\label{sec:eval-summary}

Taken together with \S\ref{sec:eval-engines}: on per-token mask latency \grid{}
leads llguidance and XGrammar at p50 on both tokenizers (Table~\ref{tab:engines}),
by more than $2\times$ over XGrammar; llguidance retains the flattest p99. The
SynCode and GBNF arms were dropped by decision (2026-07-10) as the engine
comparison had already settled on this front.

\section{Limitations and honest boundaries}\label{sec:limits}

\begin{itemize}\itemsep2pt
\item \textbf{Distribution faithfulness.} Masking changes the sampling
distribution (Remark~\ref{rem:dist}); \grid{} measures the downstream effect
(EX deltas) rather than claiming neutrality. GAD/ASAp-style faithful sampling
and CRANE-style two-phase decoding are compatible, deferred arms.
\item \textbf{Column-level RBAC is post-parse by proof}
(Proposition~\ref{prop:column-rbac}). The mask guarantee is verb- and
table-level; the checker covers the residue and feeds repair.
\item \textbf{LALR(1) languages only (v1).} MaskBench draws the boundary
concretely: 79/315 schemas are declared compile errors, 5 of them genuine LALR
conflicts. An Earley fallback is a recorded option, unexercised because no
target dialect construct has forced it.
\item \textbf{Cold-walk tails.} The cold trie walk over a 100k+ vocabulary is
milliseconds by nature; the write-back cache amortizes it across requests
(86--98\% hit rates in replay- and serving-shaped workloads) and the serving
contract keeps it off co-tenants' critical path, but one-shot workloads see it
(MaskBench p95$+$). Walk-level pruning is the named next kernel target.
\item \textbf{The cold-window serving contention.} A fresh, never-before-seen
schema induces a transient co-batched slowdown ($\sim$34\% during its
$\sim$0.66\,s first-request specialization window). The kernel-v7 fused cold-miss
path removed the GIL-bound entry-materialization cost that dominated at v6, so the
residual is genuine host CPU/memory-bandwidth contention between the cold walk and
the engine forward loop; it shrinks as walk parallelism rises and is mitigated by
scheduling niceness. The fresh request itself runs at warm speed and steady-state
co-tenant requests are unaffected. Fully eliminating it is a compute-isolation
trade-off, noted as future work (\S\ref{sec:eval-serving}).
\item \textbf{One characterized exogenous artifact.} The serving record's
per-step tail on vLLM 0.24's multiprocess topology contains a once-per-leg
frozen step that is provably not \grid{}'s (five-way exoneration,
\S\ref{sec:eval-serving}); the serving metrics use artifact-robust estimators and
the issue is filed upstream (vllm-project/vllm \#48229) rather than papered over.
\item \textbf{Termination is mode-dependent.} The full termination guarantee
(reserve-completed stops) holds in the \grid{}-owned loop; processor-only
serving weakens to EOS-only-at-ACCEPT with truncation recorded
(Proposition~\ref{prop:term}).
\item \textbf{Host honesty.} Headline performance numbers run on declared cloud
runners; local-dev numbers are labeled and used for shape, not records.
\end{itemize}

\section{Related work}\label{sec:related}

\textbf{Outlines}~\cite{willard2023efficient} reformulated guided generation as
FSM state transitions with a vocabulary index, extending to CFGs via
lexer-aware scanning over parser states---the paper whose problem statement and
formal register this work parallels, and an explicit inspiration for \grid{}'s
design. \grid{} differs in what it keys and what it promises: masks keyed on
live LALR configurations with a write-back cross-request cache rather than a
per-automaton precomputed index; policy (roles, schemas) compiled into the
language; guarantees stated with preconditions and empirically verified; an audit trail.
Notably, current Outlines releases delegate CFG constraining to
llguidance~\cite{llguidance}, so the engine-level comparison collapses into the
llguidance one (\S\ref{sec:eval-engines}).
\textbf{guidance/llguidance}~\cite{guidance,llguidance} pair an
Earley/derivative core with a token trie; llguidance keeps the flattest p99
tails and the fastest compiles in our measurements and is the bar \grid{}
names---at kernel v7 \grid{} leads it at p50 and p90 on both tokenizers
(\S\ref{sec:eval-engines}) while llguidance retains the p99 edge.
\grid{}'s answer is also architectural rather than raw-constant: configuration-keyed
write-back caching across grammar families, mask-level RBAC/schema projection,
requirement R as a committed contract, and the audit chain.
\textbf{XGrammar}~\cite{dong2024xgrammar} introduced the
context-independent/context-dependent vocabulary split with compile-time
precomputation and scheduler overlap; \grid{} adopts the CI/CD idea but makes
the split a \emph{cache-key soundness} requirement (never cache CD), computes
on miss with write-back (rewarding the per-role/per-tenant family shape), and
adds the identifier composition rule that per-schema policy needs.
\textbf{SynCode}~\cite{ugare2024syncode} formalizes grammar masks over
lexer-state $\times$ remainder tables with soundness/completeness theorems; its
shipped $d$-lookahead masks are sound but incomplete, which is incompatible
with dead-end freedom---\grid{} chooses exact byte-level masks for exactly that
reason. \textbf{PICARD}~\cite{scholak2021picard} pioneered constrained SQL
decoding by incremental parsing over beam candidates (reject-and-filter rather
than exact masks). \textbf{Grammar-aligned decoding}~\cite{park2024grammar}
and \textbf{CRANE}~\cite{banerjee2025crane} address the distribution and
reasoning costs of strict masking; both are complementary layers above an
exact-mask engine like \grid{}. Automata-theoretic treatments of the
token--terminal misalignment~\cite{koo2024automata} give the detokenization
correctness framing our byte-trie bridge instantiates.
\textbf{JSONSchemaBench}~\cite{geng2025jsonschemabench} supplies the
cross-engine schema corpus used in \S\ref{sec:eval-maskbench}. Serving
integration follows the vLLM structured-output interface~\cite{kwon2023vllm}.
Classical foundations: viable prefixes and LR
parsing~\cite{knuth1965translation,aho1974lr}; linear Earley parsing on
LR-regular grammars~\cite{leo1991general} (the recorded fallback engine).

\section{Conclusion}\label{sec:conclusion}

\grid{} shows that the properties an enterprise actually needs from constrained
decoding---policy compiled into the language, guarantees with preconditions,
position-flat guard cost, a replayable audit trail, and batch-safe serving with
honest cold-start behavior---all follow from one commitment: key every mask,
cache entry, and audit record on the parser configuration, and prove (then
verify) that the keys refine the equivalence the semantics requires. The residue
the mask provably cannot enforce is not hidden but named, checked post-parse,
and---at model scales that can use feedback---repaired. The committed record
spans engine microbenchmarks, a public schema corpus, full-dataset execution
accuracy, and an end-to-end serving benchmark, each number carried with its host
label; the remaining open costs (cold-walk tails, the once-per-schema
cold-window serving contention, one upstream engine artifact) are stated as
precisely as the wins.

\bibliographystyle{unsrt}
\bibliography{references}

\end{document}